\documentclass[10pt,twocolumn,letterpaper]{article}

\usepackage{caption}
\usepackage{titling}
\usepackage{cvpr}
\usepackage{standalone}
\usepackage{times}
\usepackage{epsfig}
\usepackage{graphicx}
\usepackage{amsmath}
\usepackage{amssymb}
\usepackage{subfigure}
\usepackage{pifont}
\usepackage{threeparttable}
\newcommand{\xmark}{\ding{55}}%
\newcommand{\tld}{\raise.17ex\hbox{$\scriptstyle\mathtt{\sim}$}}
\def\vsp{\vspace{-0.05in}}

\usepackage[pagebackref=true,breaklinks=true,letterpaper=true,colorlinks,bookmarks=false]{hyperref}

\cvprfinalcopy 

\def\cvprPaperID{3617} 

\ifcvprfinal\fi
\date{}
\begin{document}

\title{SqueezeDet: Unified, Small, Low Power Fully Convolutional Neural Networks for Real-Time Object Detection for Autonomous Driving}

\author{Bichen Wu$^1$, Alvin Wan$^1$, Forrest Iandola$^{1,2}$, Peter H. Jin$^1$, Kurt Keutzer$^{1,2}$ \\
$^1$UC Berkeley, $^2$DeepScale\\
{\tt\small bichen@berkeley.edu, alvinwan@berkeley.edu, forrest@deepscale.ai,} \\
{\tt\small phj@berkeley.edu, keutzer@berkeley.edu}
}

\maketitle


\begin{abstract}
Object detection is a crucial task for autonomous driving. In addition to
requiring high accuracy to ensure safety, object detection for autonomous
driving also requires real-time inference speed to guarantee prompt vehicle
control, as well as small model size and energy efficiency to enable
embedded system deployment. 

In this work, we propose SqueezeDet, a fully convolutional neural network for
object detection that aims to simultaneously satisfy all of the above
constraints. In our network we use convolutional layers not only to extract
feature maps but also as the output layer to compute bounding boxes and
class probabilities. The detection pipeline of our model only contains a
single forward pass of a neural network, thus it is extremely fast. Our
model is fully-convolutional, which leads to a small model size and better
energy efficiency. While achieving the same accuracy as previous baselines,
our model is 30.4x smaller, 19.7x faster, and consumes 35.2x lower energy.
The code is open-sourced at \url{https://github.com/BichenWuUCB/squeezeDet}.
\end{abstract}

\section{Introduction}
\vsp
A safe and robust autonomous driving system relies on accurate perception of the environment. To be more specific, an autonomous vehicle needs to accurately detect cars, pedestrians, cyclists, road signs, and other objects in real-time in order to make the right control decisions that ensure safety. Moreover, to be economical and widely deployable, this object detector must operate on embedded processors that dissipate far less power than powerful GPUs used for benchmarking in typical computer vision experiments.

Object detection is a crucial task for autonomous driving. Different autonomous vehicle solutions may have different combinations of perception sensors, but image based object detection is almost irreplaceable. Image sensors are cheap compared with others such as LIDAR. Image data (including video) are much more abundant than, for example, LIDAR cloud points, and are much easier to collect and annotate. Recent progress in deep learning shows a promising trend that with more and more data that cover all kinds of long-tail scenarios, we can always design more powerful neural networks with more parameters to digest the data and become more accurate and robust.

While recent research has been primarily focused on improving accuracy, for actual deployment in an autonomous vehicle, there are other issues of image object detection that are equally critical. For autonomous driving some basic requirements for image object detectors include the following: a) Accuracy. More specifically, the detector ideally should achieve $100\%$ recall with high precision on objects of interest. b) Speed. The detector should have real-time or faster inference speed to reduce the latency of the vehicle control loop. c) Small model size. As discussed in~\cite{SqueezeNet}, smaller model size brings benefits of more efficient distributed training, less communication overhead to export new models to clients through wireless update, less energy consumption and more feasible embedded system deployment. d) Energy efficiency. Desktop and rack systems may have the luxury of burning 250W of power for neural network computation, but embedded processors targeting automotive market must fit within a much smaller power and energy envelope. While precise figures vary, the new Xavier\footnote{https://blogs.nvidia.com/blog/2016/09/28/xavier/} processor from Nvidia, for example, is targeting a 20W thermal design point. Processors targeting mobile applications have an even smaller power budget and must fit in the 3W--10W range. Without addressing the problems of a) accuracy, b) speed, c) small model size, and d) energy efficiency, we won't be able to truly leverage the power of deep neural networks for autonomous driving.  

In this paper, we address the above issues by presenting \textit{SqueezeDet}, a fully convolutional neural network for object detection. The detection pipeline of SqueezeDet is inspired by~\cite{YOLO}: first, we use stacked convolution filters to extract a high dimensional, low resolution feature map for the input image. Then, we use \textit{ConvDet}, a convolutional layer to take the feature map as input and compute a large amount of object bounding boxes and predict their categories. Finally, we filter these bounding boxes to obtain final detections. The ``backbone'' convolutional neural net (CNN) architecture of our network is SqueezeNet~\cite{SqueezeNet}, which achieves AlexNet level imageNet accuracy with a model size of $<5$MB that can be further compressed to $0.5$MB. After strengthening the SqueezeNet model with additional layers followed by \textit{ConvDet}, the total model size is still less than $8$MB. The inference speed of our model can reach $57.2$ FPS\footnote{Standard camera frame rate is 30 FPS, which is regarded as the benchmark of the real-time speed.} with input image resolution of 1242x375. Benefiting from the small model size and activation size, SqueezeDet has a much smaller memory footprint and requires fewer DRAM accesses, thus it consumes only $1.4$J of energy per image on a TITAN X GPU, which is about 84X less than a Faster R-CNN model described in~\cite{ShallowNetworks}. SqueezeDet is also very accurate. One of our trained SqueezeDet models achieved the best average precision in all three difficulty levels of cyclist detection in the KITTI object detection challenge~\cite{KITTI}.

The rest of the paper is organized as follows. We first review related work in section~\ref{sec:related_work}. Then, we introduce our detection pipeline, the \textit{ConvDet} layer, the training protocol and network design of SqueezeDet in section~\ref{sec:method}. In section~\ref{sec:experiments}, we report our experiments on the KITTI dataset, and discuss accuracy, speed, parameter size of our model. Due to limited page length, we put energy efficiency discussion in the supplementary material to this paper. We conclude the paper in section~\ref{sec:conclusion}.

\section{Related Work}
\label{sec:related_work}
\vsp

\subsection{CNNs for object detection}
\vsp

From 2005 to 2013, various techniques were applied to advance the accuracy of object detection on datasets such as PASCAL~\cite{PASCAL}.
In most of these years, versions of HOG+SVM~\cite{HOG} or DPM~\cite{DPM-journal} led the state-of-art accuracy on these datasets.
However, in 2013, Girshick \etal proposed Region-based Convolutional Neural Networks (R-CNN)~\cite{R-CNN}, which led to substantial gains in object detection accuracy.
The R-CNN approach begins by identifying region proposals (i.e. regions of interest that are likely to contain objects) and then classifying these regions using a CNN.
One disadvantage of R-CNN is that it computes the CNN independently on each region proposal, leading to time-consuming ($\leq$ 1 fps) and energy-inefficient ($\geq$ 200 J/frame) computation. 
To remedy this, Girshick \etal experimented with a number of strategies to amortize computation across the region proposals~\cite{DPMareCNN,DenseNet,Fast-R-CNN}, culminating in {\em Faster R-CNN}~\cite{Faster-R-CNN}.An other model, R-FCN, is fully-convolutional and delivers accuracy that is competitive with R-CNN, but R-FCN is fully-convolutional which allows it to amortize more computation across the region proposals.

There have been a number of works that have adapted the R-CNN approach to address object detection for autonomous driving. Almost all the top-ranked published methods on the KITTI leader board are based on Faster R-CNN. \cite{ShallowNetworks} modified the CNN architecture to use shallower networks to improve accuracy. \cite{mscnn, subcnn} on the other hand focused on generating better region proposals. Most of these methods focused on better accuracy, but to our knowledge, no previous methods have reported real-time inference speeds on KITTI dataset. 

{\em Region proposals} are a cornerstone in all of the object detection methods that we have discussed so far. However, in YOLO (You Only Look Once)~\cite{YOLO}, region proposition and classification are integrated into one single stage. Compared with R-CNN and Faster R-CNN based methods, YOLO's single stage detection pipeline is extremely fast, making YOLO the first CNN based general-purpose object detection model that achieved real-time speed.

\subsection{Small CNN models}
\vsp

For any particular accuracy level on a computer vision benchmark, it is usually feasible to develop multiple CNN architectures that are able to achieve that level of accuracy.
Given the same level of accuracy, it is often beneficial to develop smaller CNNs (i.e. CNNs with fewer model parameters), as discussed in~\cite{SqueezeNet}. AlexNet~\cite{alexnet} and VGG-19~\cite{VGG-19} are CNN model architectures that were designed for image classification and have since been modified to address other computer vision tasks.
The AlexNet model contains 240MB of parameters, and it delivers approximately 80\% top-5 accuracy on ImageNet~\cite{imagenet} image classification.
The VGG-19 model contains 575MB of parameters and delivers \tld~87\% top-5 accuracy on ImageNet.
However, models with fewer parameters can deliver similar levels of accuracy.
The SqueezeNet~\cite{SqueezeNet} model has only 4.8MB of parameters (50x smaller than AlexNet), and it matches or exceeds AlexNet-level accuracy on ImageNet.
The GoogLeNet-v1~\cite{googlenet} model only has 53MB of parameters, and it matches VGG-19-level accuracy on ImageNet.

\subsection{Fully convolutional networks}
\vsp

{\em Fully-convolutional networks (FCN)} were popularized by Long \etal, who applied them to the semantic segmentation domain~\cite{FCN}.
FCN defines a broad class of CNNs, where the output of the final parameterized layer is a grid rather than a vector.\footnote{By ``parameterized layer," we are referring to layers (e.g. convolution or fully-connected) that have parameters that are learned from data. Pooling or ReLU layers are not parameterized layers because they have no learned parameters.} 
This is useful in semantic segmentation, where each location in the grid corresponds to the predicted class of a pixel. 

FCN models have been applied in other areas as well.
To address the image classification problem, a CNN needs to output a 1-dimensional vector of class probabilities.
One common approach is to have one or more {\em fully-connected layers}, which by definition output a 1D vector -- 1$\times$1$\times$Channels (e.g.~\cite{alexnet,VGG-19}).
However, an alternative approach is to have the final parameterized layer be a convolutional layer that outputs a grid (H$\times$W$\times$Channels), and to then use average-pooling to downsample the grid to 1$\times$1$\times$Channels to a vector of produce class probabilities (e.g.~\cite{SqueezeNet,NiN}).
Finally, the R-FCN method that we mentioned earlier in this section is a fully-convolutional network. 

\section{Method Description}
\label{sec:method}
\vsp

\subsection{Detection Pipeline}
\begin{figure}[h]
  \centering
 \includegraphics [width=3in]{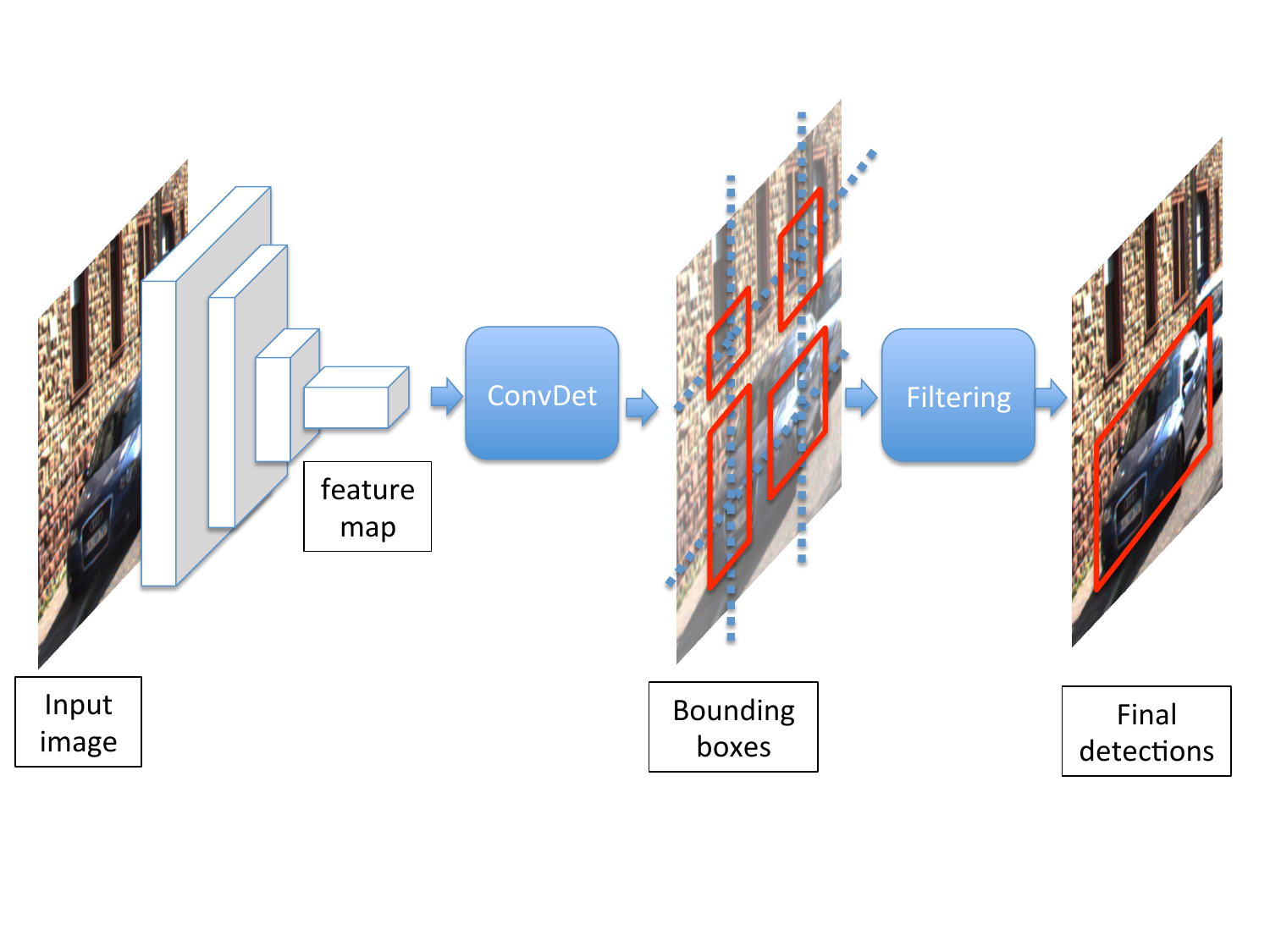}
 \caption{SqueezeDet detection pipeline. A convolutional neural network extracts a feature map from the input image and feeds it into the \textit{ConvDet} layer. The \textit{ConvDet} layer then computes bounding boxes centered around $W\times H$ uniformly distributed grid centers. Each bounding box is associated with $1$ confidence score and $C$ conditional class probabilities. Then, we keep the top $N$ bouding boxes with highest confidence and use NMS to filter them to get the final detections.}
\label{fig:DetPipeline}
\end{figure}

\vsp
Inspired by YOLO~\cite{YOLO}, we adopt a single-stage detection pipeline: region proposition and classification is performed by one single network simultaneously. As shown in Fig.\ref{fig:DetPipeline}, a convolutional neural network first takes an image as input and extract a low-resolution, high dimensional feature map. Then, the feature map is fed it into the \textit{ConvDet} layer to compute bounding boxes centered around $W\times H$ uniformly distributed spatial grids. $W$ and $H$ are the number of grid centers along horizontal and vertical directions. 

\vsp

Each bounding box is associated with $C+1$ values, where $C$ is the number of classes to distinguish, and the extra $1$ is for the confidence score, which indicates how likely does the bounding box contain an object. Similarly to YOLO~\cite{YOLO}, we define the confidence score as $Pr(\text{Object})*\text{IOU}_{truth}^{pred}$. A high confidence score implies a high probability that an object of interest does exist and that the overlap between the predicted bounding box and the ground truth is high. The other $C$ scalars represents the conditional class probability distribution given that the object exists within the bounding box. More formally, we denote the conditional probabilities as $Pr(\text{class}_c|\text{Object}), c \in [1,C].$ We assign the label with the highest conditional probability to this bounding box and we use 
\[
	\max_c Pr(\text{class}_c|\text{Object}) * Pr(\text{Object})*\text{IOU}_{truth}^{pred}
\]
as the metric to estimate the confidence of the bounding box prediction. 

Finally, we keep the top $N$ bounding boxes with the highest confidence and use Non-Maximum Suppression (NMS) to filter redundant bounding boxes to obtain the final detections. During inference, the entire detection pipeline consists of only one forward pass of one neural network with minimal post-processing.

\subsection{ConvDet}
\label{sec:convdet}
\vsp

The \textit{SqueezeDet} detection pipeline is inspired by YOLO~\cite{YOLO}. But as we will describe in this section, the design of the \textit{ConvDet} layer enables SqueezeDet to generate tens-of-thousands of region proposals with much fewer model parameters compared to YOLO. 

\textit{ConvDet} is essentially a convolutional layer that is trained to output bounding box coordinates and class probabilities. It works as a sliding window that moves through each spatial position on the feature map. At each position, it computes \(K\times(4+1+C)\) values that encode the bounding box predictions. Here, $K$ is the number of reference bounding boxes with pre-selected shapes. Using the notation from~\cite{Faster-R-CNN}, we call these reference bounding boxes as anchor. Each position on the feature map corresponds to a grid center in the original image, so each anchor can be described by 4 scalars as \( (\hat{x}_i, \hat{y}_j, \hat{w}_k, \hat{h}_k), i \in [1, W], j\in [1, H], k \in [1, K]. \) Here $\hat{x}_i, \hat{y}_i$ are spatial coordinates of the reference grid center $(i, j)$. $\hat{w}_k, \hat{h}_k$ are the width and height of the $k$-th reference bounding box. We used the method described by~\cite{ShallowNetworks} to select reference bounding box shapes to match the data distribution. 

\begin{figure}[h]
  \centering
 \includegraphics [width=3in]{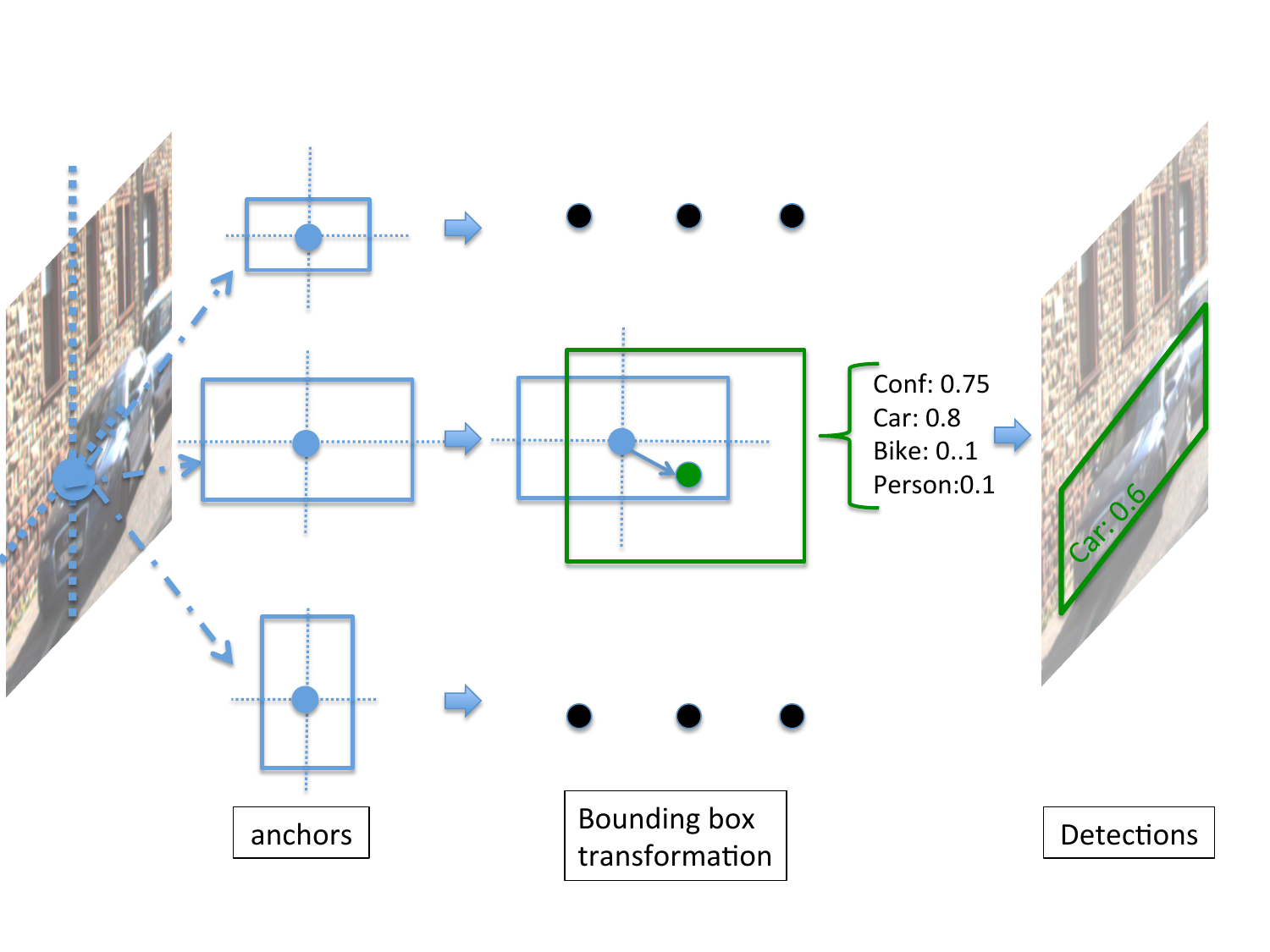}
 \caption{Bounding box transformation. Each grid center has $K$ anchors with pre-selected shapes. Each anchor is transformed to its new position and shape using the relative coordinates computed by the \textit{ConvDet} layer. Each anchor is associated with a confidence score and class probabilities to predict the category of the object within the bounding box.}
\label{fig:bbox_transform}
\end{figure}

For each anchor $(i, j, k)$, we compute $4$ relative coordinates \((\delta x_{ijk},  \delta y_{ijk}, \delta w_{ijk}, \delta h_{ijk})\) to transform the anchor into a predicted bounding box, as shown in Fig. \ref{fig:bbox_transform}. Following~\cite{R-CNN-supp}, the transformation is described by 
\begin{equation}
\label{eq:bbox_trans}
\begin{gathered}
	x_i^p = \hat{x}_i + \hat{w}_k\delta x_{ijk}, \\
    y_j^p = \hat{y}_j + \hat{h}_k\delta y_{ijk}, \\
    w_k^p = \hat{w}_k \exp(\delta w_{ijk}), \\
    h_k^p = \hat{h}_k \exp(\delta h_{ijk}),
\end{gathered}
\end{equation}
where $x_i^p, y_j^p, w_k^p, h_k^p$ are predicted bounding box coordinates. As explained in the previous section, the other $C+1$ outputs for each anchor encode the confidence score for this prediction and conditional class probabilities.

\vsp

\textit{ConvDet} is similar to the last layer of RPN in Faster R-CNN~\cite{Faster-R-CNN}. The major difference is that, RPN is regarded as a ``weak'' detector that is only responsible for detecting whether an object exists and generating bounding box proposals for the object. The classification is handed over to fully connected layers, which are regarded as a ``strong'' classifier.In fact, convolutional layers are ``strong'' enough to detect, localize, and classify objects at the same time.

\begin{figure}[h]
\hfill
\centering
\subfigure[Last layer of Region Proposal Network (RPN) is a 1x1 convolution with $K\times (4+1)$ outputs. $4$ is the number of relative coordinates, and $1$ is the confidence score. It's only responsible for generating region proposals. The parameter size for this layer is $\text{Ch}_f \times K \times 5$.]{ \includegraphics[clip, trim=1cm 3.5cm 2.5cm 6cm, width=1\linewidth]{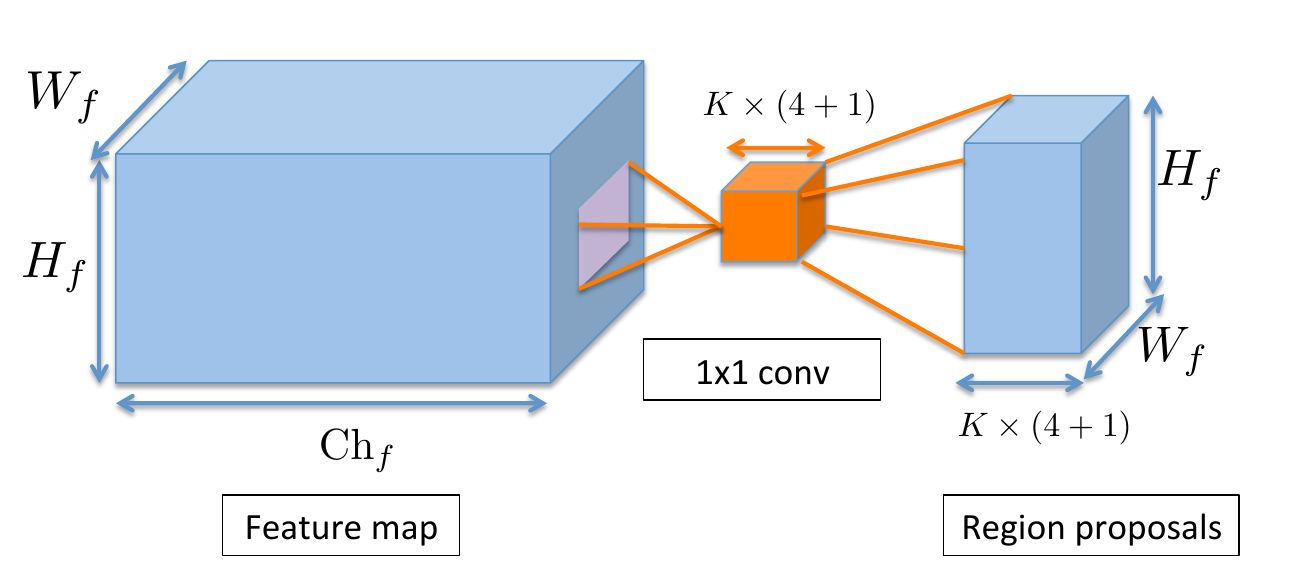}}
\hfill
\subfigure[The \textit{ConvDet} layer is a $F_w \times F_h$ convolution with output size of $K\times (5+C)$. It's responsible for both computing bounding boxes and classifying the object within. The parameter size for this layer is $F_w F_h \text{Ch}_fK(5+C)$. ]{ \includegraphics[clip, trim=1cm 3.5cm 1cm 5cm, width=1\linewidth]{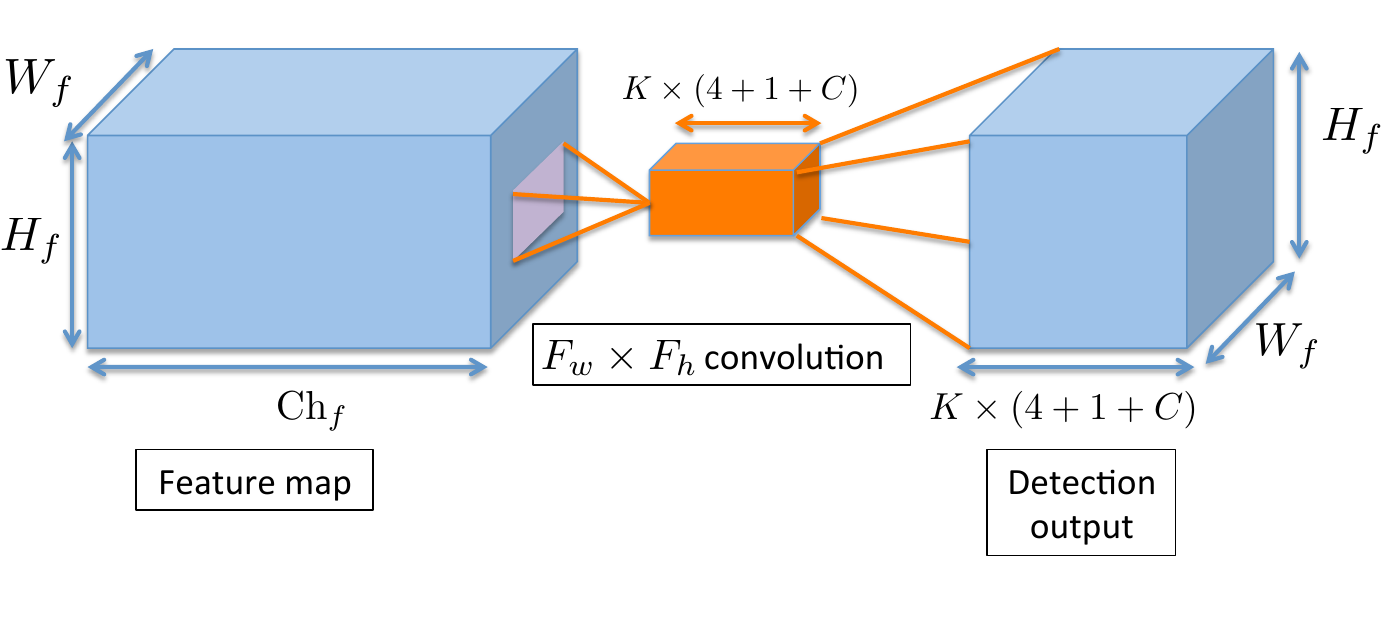}}
\hfill
\subfigure[The detection layer of YOLO~\cite{YOLO} contains 2 fully connected layers. The first one is of size $W_fH_f\text{Ch}_fF_{fc1}$. The second one is of size $F_{fc1} W_oH_oK(5+C)$.]{ \includegraphics[clip, trim=0cm 5cm 0cm 1cm, width=1\linewidth]{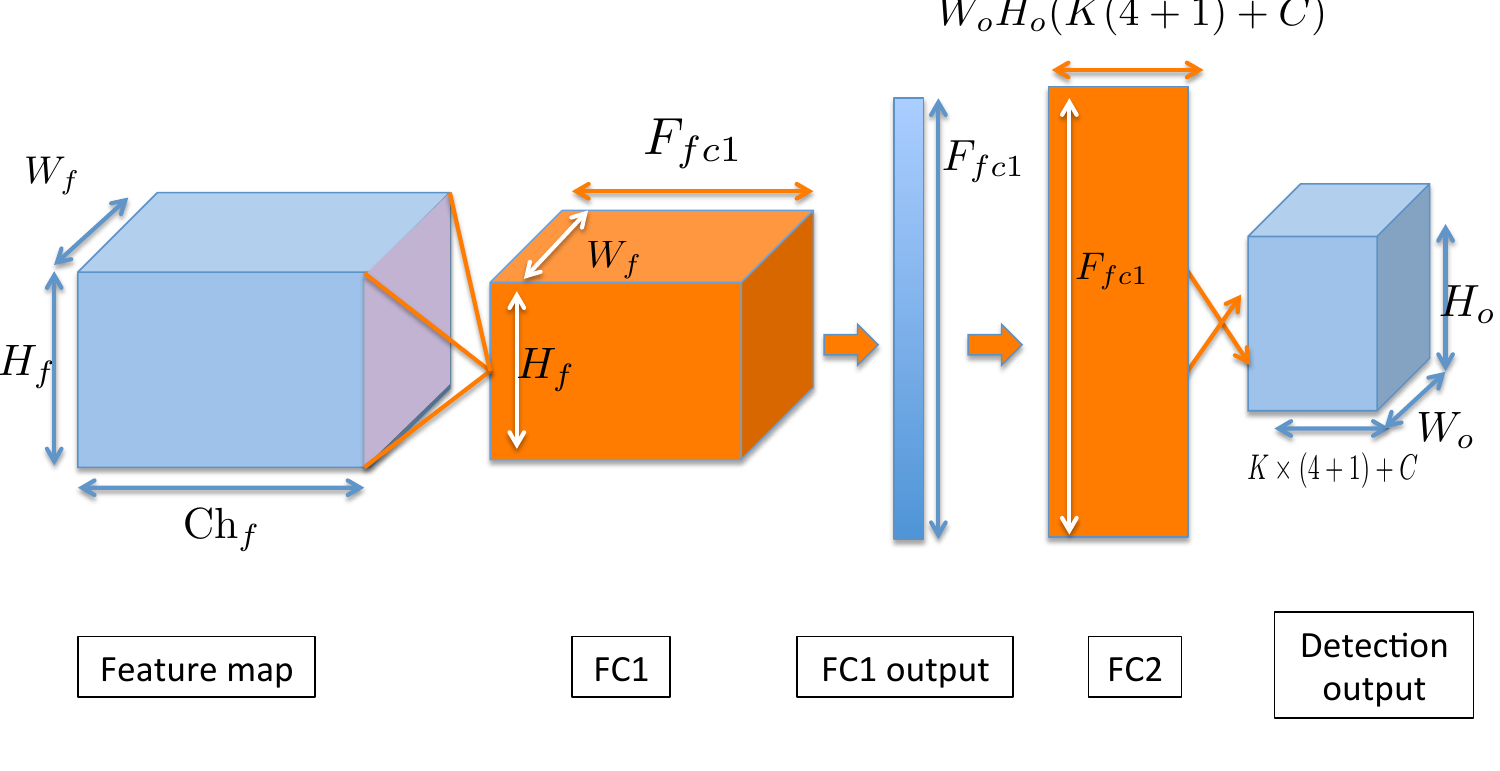}}
\hfill
\vspace{-1pt}
\caption{Comparing RPN, \textit{ConvDet} and the detection layer of YOLO~\cite{YOLO}. Activations are represented as blue cubes and layers (and their parameters) are represented as orange ones. Activation and parameter dimensions are also annotated.}
\label{fig:RPN_ConvDet_YOLO}
\end{figure}

For simplicity, we denote the detection layers of YOLO~\cite{YOLO} as \textit{FcDet} (only counting the last two fully connected layers). Compared with \textit{FcDet}, the \textit{ConvDet} layer has orders of magnitude fewer parameters and is still able to generate more region proposals with higher spatial resolution. The comparison between \textit{ConvDet} and \textit{FcDet} is illustrated in Fig.~\ref{fig:RPN_ConvDet_YOLO}. 

Assume that the input feature map is of size $(W_f, H_f, \text{Ch}_f)$, $W_f$ is the width of the feature map, $H_f$ is the height, and $\text{Ch}_f$ is the number of input channels to the detection layer. Denote \textit{ConvDet}'s filter width as $F_w$ and height as $F_h$. With proper padding/striding strategy, the output of \textit{ConvDet} keeps the same spatial dimension as the feature map. To compute $K\times(4+1+C)$ outputs for each reference grid, the number of parameters required by the \textit{ConvDet} layer is $F_wF_h\text{Ch}_fK(5+C)$. 

The \textit{FcDet} layer described in \cite{YOLO} is comprised of two fully connected layers. Using the same notation for the input feature map and assuming the number of outputs of the $fc1$ layer is $F_{fc1}$, then the number of parameters in the $fc1$ layer is $W_fH_f\text{Ch}_fF_{fc1}$. The second fully connected layer in \cite{YOLO} generates $C$ class probabilities as well as $K\times(4+1)$ bounding box coordinates and confidence scores for each of the $W_o \times H_o$ grids. Thus, the number of parameters in the $fc2$ layer is $F_{fc1}W_oH_o(5K+C)$. The total number of parameters in these two fully connected layers is $F_{fc1}(W_fH_f\text{Ch}_f + W_oH_o(5K+C))$. 

In \cite{YOLO}, the input feature map is of size 7x7x1024. $F_{fc1} = 4096$, $K=2$, $C=20$, $W_o = H_o = 7$, thus the total number of parameters required by the two fully connected layers is approximately $212\times 10^6$. If we keep the feature map sizes, number of output grid centers, classes, and anchors the same, and use 3x3 \textit{ConvDet}, it would only require $3\times3\times 1024\times 2 \times 25 \approx 0.46\times 10^6$ parameters, 460X smaller then \textit{FcDet}. The comparison of RPN, \textit{ConvDet} and \textit{FcDet} is illustrated in Fig.~\ref{fig:RPN_ConvDet_YOLO} and Table~\ref{table:RPN_CONVDET_YOLO}.

\begin{table}
\begin{center}
\begin{tabular}{c|ccc}

 & RP & $cls$ & \#Parameter\\
\hline
RPN & \checkmark & \xmark & $5\text{Ch}_fK$ \\
\textit{ConvDet} & \checkmark & \checkmark & $F_wF_h\text{Ch}_fK(5+C)$ \\
\textit{FcDet} & \checkmark & \checkmark & $F_{fc1}(W_fH_f\text{Ch}_f + W_oH_o(5K+C))$\\
\hline
\end{tabular}
\end{center}
\caption{Comparison between RPN, \textit{ConvDet} and \textit{FcDet}. RP stands for region proposition. $cls$ stands for classification. }
\label{table:RPN_CONVDET_YOLO}
\end{table}

\vsp

\subsection{Training protocol}
\vsp

Unlike Faster R-CNN~\cite{Faster-R-CNN}, which deploys a (4-step) alternating training strategy to train RPN and detector network, our SqueezeDet detection network can be trained end-to-end, similarly to YOLO~\cite{YOLO}.

To train the \textit{ConvDet} layer to learn detection, localization and classification, we define a multi-task loss function:   
\begin{equation}
\label{eq:loss}
\begin{gathered}
	\frac{\lambda_{bbox}}{N_{obj}} \sum_{i=1}^W \sum_{j=1}^H \sum_{k=1}^K 
    I_{ijk} [(\delta x_{ijk} - \delta x_{ijk}^G)^2
    +(\delta y_{ijk} - \delta y_{ijk}^G)^2 \\ 
    + (\delta w_{ijk} - \delta w_{ijk}^G)^2 + (\delta h_{ijk} - \delta h_{ijk}^G)^2]\\
    +  \sum_{i=1}^W \sum_{j=1}^H \sum_{k=1}^K 
   \frac{\lambda_{conf}^+}{N_{obj}} I_{ijk} (\gamma_{ijk} - \gamma_{ijk}^G)^2 
   + \frac{\lambda_{conf}^-}{WHK - N_{obj}}\bar{I}_{ijk} \gamma_{ijk}^2 \\
   + \frac{1}{N_{obj}} \sum_{i=1}^W \sum_{j=1}^H \sum_{k=1}^K \sum_{c=1}^C
   I_{ijk} l_c^G \log(p_c).
\end{gathered}
\end{equation}

The first part of the loss function is the bounding box regression. \((\delta x_{ijk}, \delta y_{ijk}, \delta w_{ijk}, \delta h_{ijk})\) corresponds to the relative coordinates of anchor-$k$ located at grid center-$(i,j)$. They are outputs of the \textit{ConvDet} layer. The ground truth bounding box $\delta_{ijk}^G$, or \((\delta x_{ijk}^G, \delta y_{ijk}^G, \delta w_{ijk}^G, \delta h_{ijk}^G)\), is computed as: 
\begin{equation}
\label{eq:bbox_trans_inv}
\begin{gathered}
	\delta x_{ijk}^G = (x^G - \hat{x}_i)/\hat{w}_k, \\
    \delta y_{ijk}^G = (y^G - \hat{y}_j)/\hat{h}_k, \\
    \delta w_{ijk}^G = \log(w^G/\hat{w}_k), \\
    \delta h_{ijk}^G = \log(h^G/\hat{h}_k). \\
\end{gathered}
\end{equation}
Note that Equation~\ref{eq:bbox_trans_inv} is essentially the inverse transformation of Equation~\ref{eq:bbox_trans}. \((x^G, y^G, w^G, h^G)\) are coordinates of a ground truth bounding box. During training, we compare ground truth bounding boxes with all anchors and assign them to the anchors that have the largest overlap (IOU) with each of them. The reason is that we want to select the ``closest'' anchor to match the ground truth box such that the transformation needed is reduced to minimum. $I_{ijk}$ evaluates to 1 if the $k$-th anchor at position-$(i, j)$ has the largest overlap with a ground truth box, and to 0 if no ground truth is assigned to it. This way, we only include the loss generated by the ``responsible'' anchors. As there can be multiple objects per image, we normalize the loss by dividing it by the number of objects.

The second part of the loss function is confidence score regression. \(\gamma_{ijk}\) is the output from the \textit{ConvDet} layer, representing the predicted confidence score for anchor-$k$ at position-$(i,j)$.  \(\gamma_{ijk}^G\) is obtained by computing the IOU of the predicted bounding box with the ground truth bounding box. Same as above, we only include the loss generated by the anchor box with the largest overlap with the ground truth. For anchors that are not ``responsible'' for the detection, we penalize their confidence scores with the $\bar{I}_{ijk}\gamma_{ijk}^2$ term, where $\bar{I}_{ijk} = 1 - I_{ijk}$. Usually, there are much more anchors that are not assigned to any object. In order to balance their influence, we use $\lambda_{conf}^+$ and $\lambda_{conf}^-$ to adjust the weight of these two loss components. By definition, the confidence score's range is [0, 1]. To guarantee that $\gamma_{ijk}$ falls into that range, we feed the corresponding \textit{ConvDet} output into a \textit{sigmoid} function to normalize it.

The last part of the loss function is just cross-entropy loss for classification. $l_c^G \in \{0, 1\}$ is the ground truth label and $p_c \in [0, 1], c\in[1, C]$ is the probability distribution predicted by the neural net. We used \textit{softmax} to normalize the corresponding \textit{ConvDet} output to make sure that $p_c$ is ranged between $[0, 1]$.

The hyper-parameters in Equation~\ref{eq:loss} are selected empirically. In our experiments, we set $\lambda_{bbox}=5, \lambda_{conf}^+=75,  \lambda_{conf}^-=100$. This loss function can be optimized directly using back-propagation.  

\subsection{Neural Network Design}
\vsp

So far in this section, we described the single-stage detection pipeline, the \textit{ConvDet} layer, and the end-to-end training protocol. These parts are universal and can work with various CNN architectures, including VGG16\cite{VGG}, ResNet\cite{resnet}, etc. When choosing the ``backbone'' CNN structure, our focus is mainly on model size and energy efficiency, and SqueezeNet\cite{SqueezeNet} is our top candidate. 

\textbf{Model size.} SqueezeNet is built upon \textit{Fire Module}, which is comprised of a \textit{squeeze} layer as input, and two parallel \textit{expand} layers as output. The \textit{squeeze} layer is a 1x1 convolutional layer that compresses an input tensor with large channel size to one with the same batch and spatial dimension, but smaller channel size. The \textit{expand} layer is a mixture of 1x1 and 3x3 convolution filters that takes the compressed tensor as input, retrieve the rich features and output an activation tensor with large channel size. The alternating \textit{squeeze} and \textit{expand} layers effectively reduces parameter size without losing too much accuracy.

\textbf{Energy efficiency.} Different operations involved in neural network inference have varying energy needs. The most expensive operation is DRAM access, which uses 100 times more energy than SRAM access and floating point operations~\cite{han2016eie}. Thus, we want to reduce DRAM access as much as possible. 

The most straightforward strategy to reduce DRAM access is to use small models which reduces memory access for parameters. An effective way to reduce parameter size is to use convolutional layers instead of fully connected layers when possible. Convolution parameters can be accessed once and reused across all neighborhoods of all data items (if batch$>$1) of the input data. However, the FC layer only exposes parameter reuse opportunities in the ``batch" dimension, and each parameter is only used on one neighborhood of the input data. Besides model size, another important aspect is to control the size of intermediate activations. Assume the SRAM size of the computing hardware is 16MB, the SqueezeNet model size is 5MB. If the total size of activation output of any two consecutive layers is less than 11MB, then all the memory accesses can be completed in SRAM, no DRAM accesses are needed. A detailed energy efficiency discussion will be provided as supplementary material to this paper. 

In this paper, we adopted two versions of the SqueezeNet architecture. The first one is the SqueezeNet v1.1 model\footnote{\url{https://github.com/DeepScale/SqueezeNet/}}
with $4.72$MB of model size and $>80.3\%$ ImageNet top-5 accuracy. The second one is a more powerful SqueezeNet variation with squeeze ratio of $0.75$, $86.0\%$ of ImageNet accuracy and $19$MB of model size~\cite{SqueezeNet}. In this paper, we denote the first model as SqueezeDet and the second one as SqueezeDet+. We pre-train these two models for ImageNet classification, then we add two fire modules with randomly initialized weight on top of the pretrained model, and connect to the \textit{ConvDet} layer. 

\begin{table*}[h!]
\centering
\begin{tabular}{c|cccccc}
Method                                           & \begin{tabular}[c]{@{}c@{}}Car\\ mAP \end{tabular} & \begin{tabular}[c]{@{}c@{}}Cyclist\\ mAP \end{tabular} & \begin{tabular}[c]{@{}c@{}}Pedestrian\\ mAP \end{tabular} & \begin{tabular}[c]{@{}c@{}}All\\ mAP \end{tabular} & \begin{tabular}[c]{@{}c@{}}Model size\\ (MB)\end{tabular} & \begin{tabular}[c]{@{}c@{}}Speed\\ (FPS)\end{tabular} \\ \hline
FRCN + VGG16\cite{ShallowNetworks}  & 86.0                                                   & -                                                          & -                                                             & -                                                      & 485                                                       & 1.7                                                   \\
FRCN + AlexNet\cite{ShallowNetworks} & 82.6                                                   & -                                                          & -                                                             & -                                                      & 240                                                       & 2.9                                                   \\ \hline
SqueezeDet   (ours)                                    & 82.9                                                   & 76.8                                                       & 70.4                                                          & 76.7                                                   & \textbf{7.9}                                              & \textbf{57.2}                                         \\
SqueezeDet+   (ours)                                   & 85.5                                                   & \textbf{82.0}                                              & \textbf{73.7}                                                 & \textbf{80.4}                                          & 26.8                                                      & 32.1                                                  \\
VGG16-Det   (ours)                                     & \textbf{86.9}                                          & 79.6                                                       & 70.7                                                          & 79.1                                                   & 57.4                                                      & 16.6                                                  \\
ResNet50-Det  (ours)                                   & 86.7                                                   & 80.0                                                       & 61.5                                                          & 76.1                                                   & 35.1                                                      & 22.5                                                  \\ \hline
\end{tabular}
\caption{Summary of detection accuracy, model size, and inference speed. The mAP (mean-average precision) for each category are averaged across three difficulty levels. The mAP for All is averaged across all categories and difficulty levels. }
\label{table:AP}
\end{table*}

\begin{table*}[h!]
\footnotesize
\begin{center}
\begin{tabular}{c|ccc|ccc|ccc|c}
& \multicolumn{3}{c|}{car} & \multicolumn{3}{c|}{cyclist} &\multicolumn{3}{c|}{pedestrian} & mAP \\
method & E & M & H & E & M & H & E & M & H & \\
\hline
FRCN + VGG16\cite{ShallowNetworks} & 92.9 & 87.9 & 77.3 & - & - & - & - & - & - & - \\
FRCN + AlexNet\cite{ShallowNetworks} & \textbf{94.7} & 84.8 & 68.3 & - & - & - & - & - & - & - \\
\hline
SqueezeDet & 90.2 & 84.7 & 73.9 & 82.9 & 75.4 & 72.1 & 77.1 & 68.3 & 65.8 &76.7 \\
SqueezeDet+ & 90.4 & 87.1 & 78.9 & \textbf{87.6} & \textbf{80.3} & \textbf{78.1} & \textbf{81.4} & \textbf{71.3} & \textbf{68.5} & \textbf{80.4} \\
VGG16-Det & 93.5 & 88.1 & 79.2 & 85.2 & 78.4 & 75.2 & 77.9 & 69.1 & 65.1 & 79.1 \\
ResNet50-Det & 92.9 & \textbf{87.9} & \textbf{79.4} & 85.0 & 78.5 & 76.6 & 67.3 & 61.6 & 55.6 & 76.1 \\
\hline
\end{tabular}
\end{center}
\caption{Detailed average precision result for each difficulty level and category.}
\label{tab:detailed-AP}
\end{table*}

\section{Experiments}
\label{sec:experiments}

We evaluated the model on the KITTI~\cite{KITTI} object detection dataset, which is designed with autonomous driving in mind. We analyzed our model's accuracy measured by average precision (AP), recall, speed, and model size, and then compare with our previous work \cite{ShallowNetworks}, a faster-RCNN based object detector trained on the KITTI dataset under the same experimental setting. Next, we analyzed the trade-off between accuracy and cost in terms of model size, FLOPS, and activation size by tuning several key hyperparameters. We implemented training, evaluation, error analysis, and visualization pipeline using Tensorflow~\cite{TensorFlow}, compiled with the cuDNN~\cite{cuDNN} computational kernels. The code is open-sourced. The energy efficiency experiments of our model will be reported in the supplementary material.

\subsection{KITTI object detection}
\vsp

\textbf{Experimental setup.} 
In our experiments, unless specified otherwise, we scaled all the input images to 1242x375. We randomly split the $7381$ training images in half into a training set and a validation set. SqueezeDet, including variations, and the baseline models \cite{ShallowNetworks} are trained and evaluated on the same training-validation dataset. Our average precision (AP) results are from the validation set. We used Stochastic Gradient Descent with momentum to optimize the loss function. We set the initial learning rate to $0.01$, learning rate decay factor to $0.5$ and decay step size to $10000$. Instead of using a fixed number of steps, we trained our model all the way until the mean average precision (mAP)\footnote{Mean of average precision in 3 difficulty levels (easy, medium, hard) of 3 categories (car, cyclist, pedestrian).} on the training set converges, and then evaluate the model on the validation set. Unless otherwise specified, we used a batch size of 20. We adopted data augmentation techniques such as random cropping and flipping to reduce overfitting.  We trained our model to detect three categories of object, car, cyclist, pedestrian. We used 9 anchors for each grid in our model. At the inference stage, we only kept the top 64 detections with the highest confidence, and use NMS to filter the bounding boxes. We used NVIDIA TITAN X GPUs for our experiments.  

\begin{figure*}[h]
\hfill
\centering
\subfigure[Example of a background error. The detector is confused by a car mirrored in the window.]{ \includegraphics[clip, trim=0cm 6cm 0cm 6cm, width=.4\linewidth]{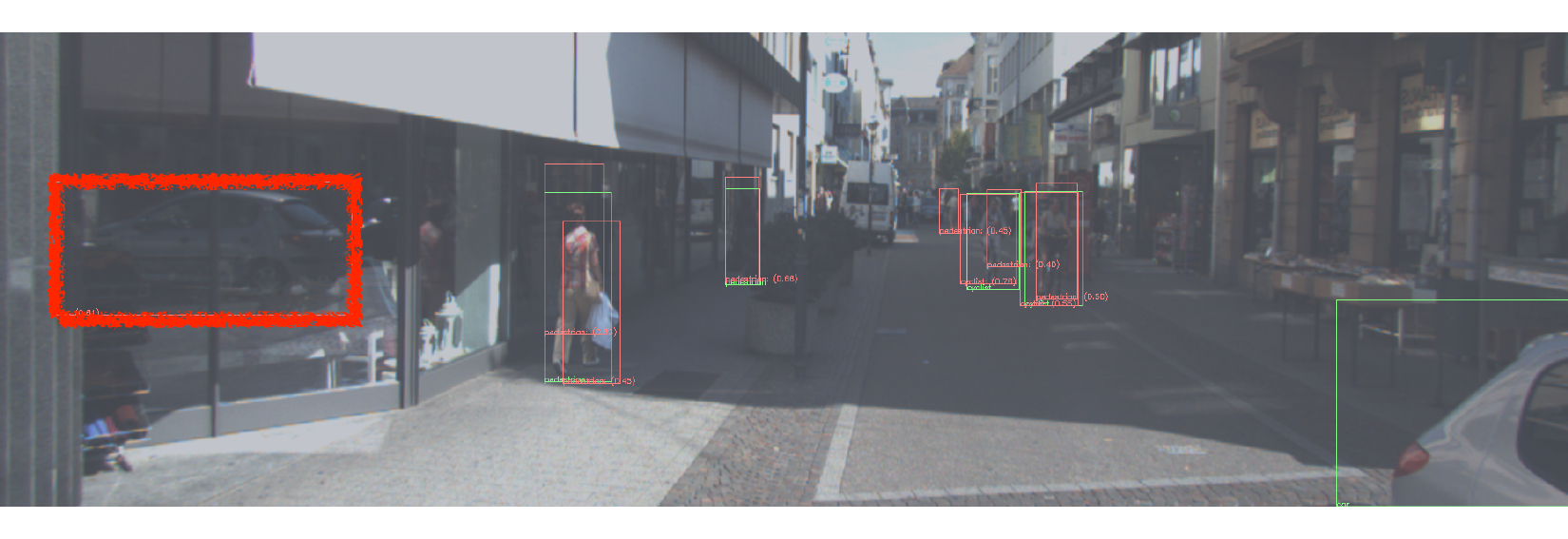}}
\subfigure[Classification error. The detector predict a cyclist to be a pedestrian.]{ \includegraphics[clip, trim=0cm 6cm 0cm 6cm, width=.4\linewidth]{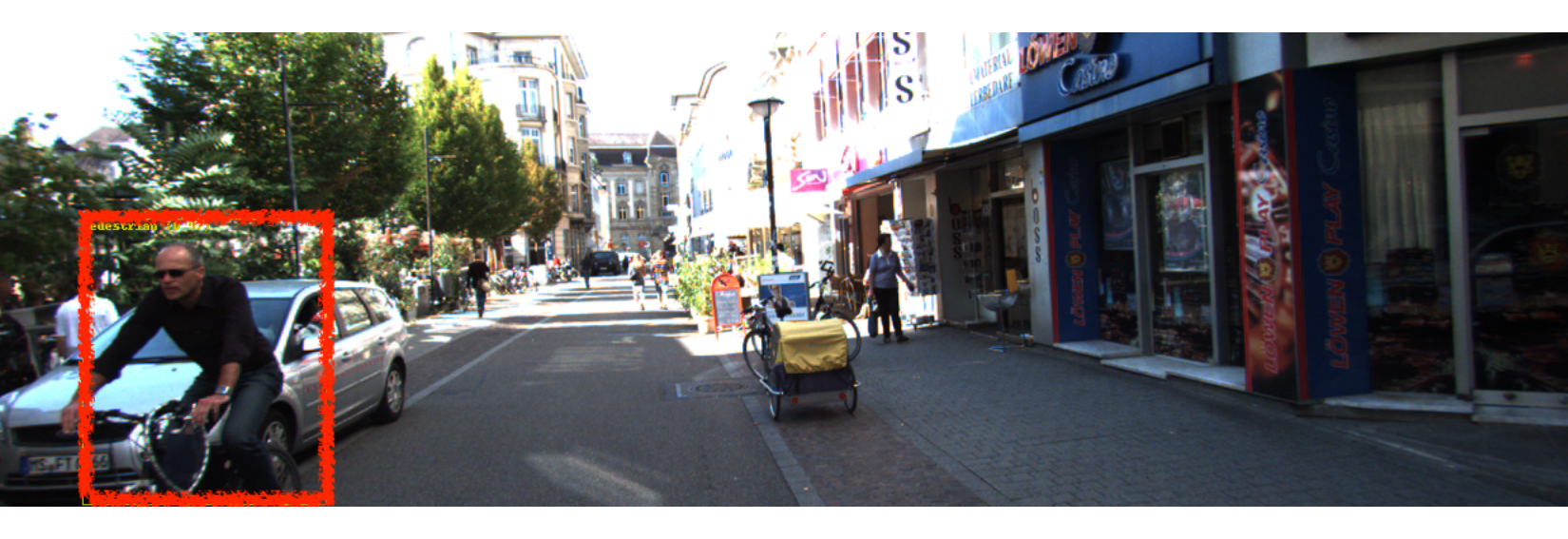}}
\hfill
\vspace{-10pt}
\subfigure[Localization error. The predicted bounding box doesn't have an IOU $>0.7$ with the ground truth.]{ \includegraphics[clip, trim=0cm 6cm 0cm 6cm, width=.4\linewidth]{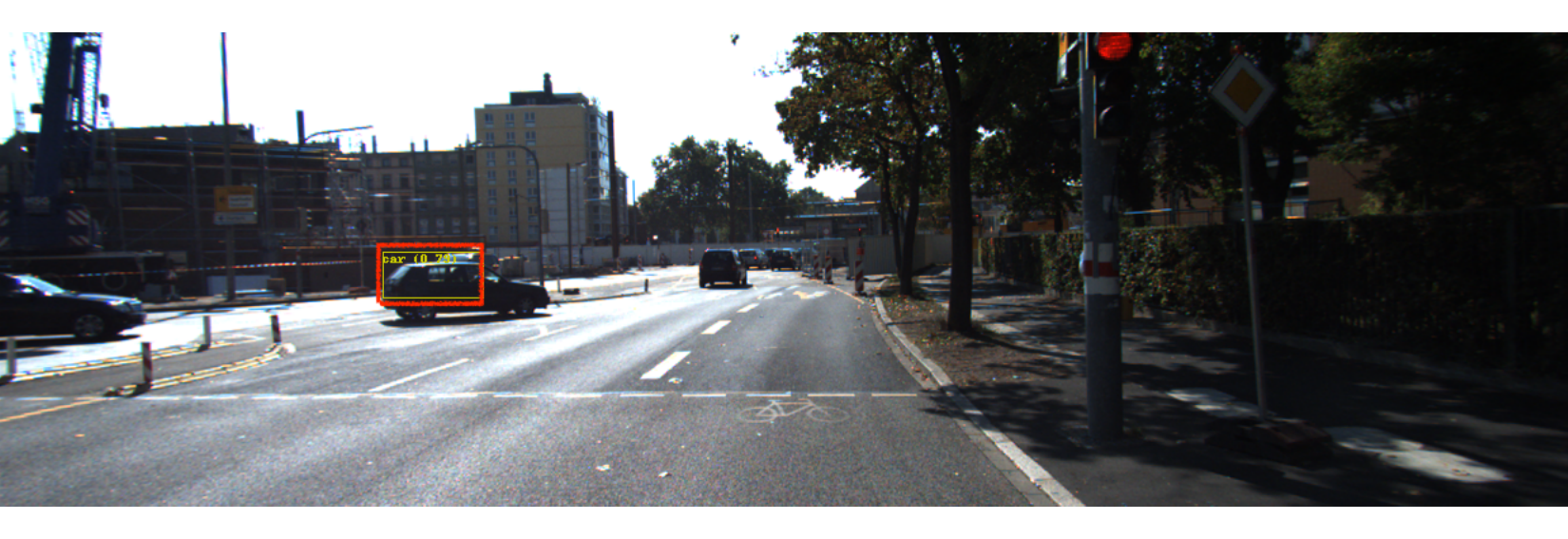}}
\subfigure[Missed object. The missed car is highly truncated and overlapped with other cars.]{ \includegraphics[clip, trim=0cm 6cm 0cm 6cm, width=.4\linewidth]{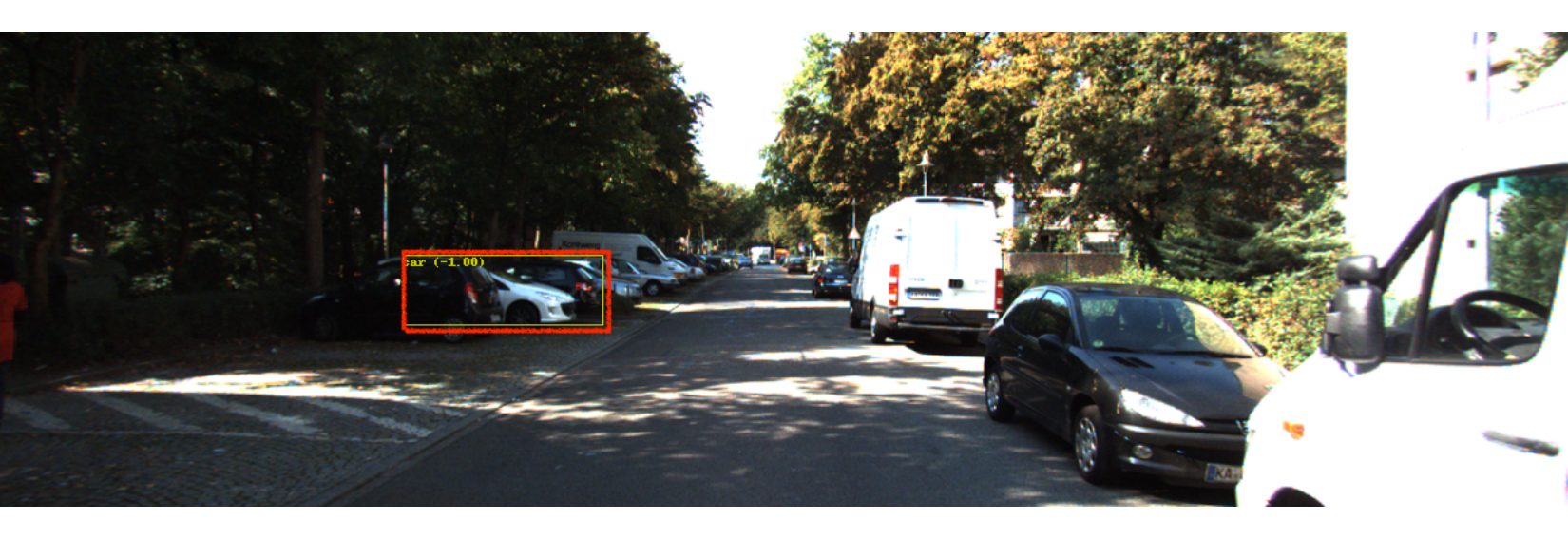}}
\caption{Example of detection errors.}
\label{fig:det_samples}
\end{figure*}

\textbf{Average Precision.} 
The detection accuracy, measured by average precision, is shown in Table~\ref{table:AP}. Our proposed SqueezeDet+ model achieved the highest mean average precision among all classes and difficulty levels. Compared with the baseline \cite{ShallowNetworks}, SqueezeDet+ is on-par with the Faster-RCNN + VGG16 model in terms of car detection accuracy. To evaluate whether \textit{ConvDet} can be applied to other backbone CNNs, we appended \textit{ConvDet} to the convolution layers of the VGG16 and ResNet50 models. Both variations achieved competitive AP. Example of error detections of SqueezeDet by types are visualized in Fig.~\ref{fig:det_samples}. More detailed accuracy results are reported in Table~\ref{tab:detailed-AP}.

\begin{figure}[h]
  \centering
  \includegraphics[clip, trim=3cm 5cm 6.5cm 5cm, width=3in]{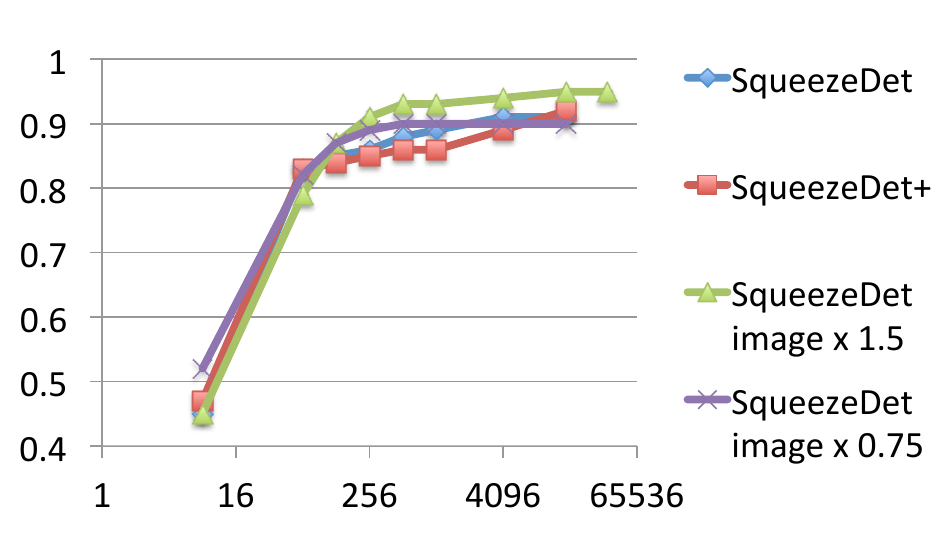}
 \caption{Overall recall vs $N_{obj}$ for SqueezeDet and SqueezeDet+ models. We also tried to re-scale the input image by 1.5X and 0.75X. SqueezeDet and SqueezeDet+ model achieved the best recall of 0.91 and 0.92 with all bounding boxes. SqueezeDet with 1.5X image resolution achieved 0.95. SqueezeDet with 0.75X image resolution achieved 0.90.}
\label{fig:recall}
\end{figure}

\textbf{Recall.} Recall is an essential metric for the safety of autonomous vehicles, so we now analyze the recall of our proposed models. For each image with a resolution of 1242x375, SqueezeDet generates in total 15048 bounding box predictions. It is intractable to perform non-maximum suppression on this many bounding boxes because of the quadratic time complexity of NMS with respect to the number of bounding boxes. Thus we only kept the top 64 predictions to feed into NMS. An interesting question to ask is, how does the number of bounding boxes kept affect recall? We tested this with the following experiment: First, we collect all the bounding box predictions and sort them by their confidence. Next, for each image, we choose the top $N_{box}$ bounding box predictions, and sweep $N_{box}$ from 8 to 15048. Then, we evaluate the overall recall for all difficulty levels of all categories. The Recall-$N_{box}$ curve is plotted in Fig.~\ref{fig:recall}. As we could see, for SqueezeDet and its strengthened model, the top 64 bounding boxes' overall recall is already larger than 80\%. If using all the bounding boxes, the SqueezeDet models can achieve $91\%$ and $92\%$ overall recall. Increasing the image size by 1.5X, the total number of bounding boxes increased to $35,190$, and the maximum recall using all bounding boxes increases to 95\%.

\textbf{Speed.} We benchmark the inference speed of SqueezeDet and baselines on a TITAN X GPU with a batch size of 1. Our models are the first to achieve real-time inference speed on KITTI dataset. Compared with the baseline \cite{ShallowNetworks}, SqueezeDet+ model achieved almost the same accuracy as Faster-RCNN+VGG16, but the inference speed is 19x faster. The smaller SqueezeDet achieved a speed of 57.2 frames per second, which is almost twice the standard of real-time speed (30 FPS).

\begin{figure}[h]
\centering
  \includegraphics[clip, trim=1cm 3cm 0cm 3cm, width=1\linewidth]{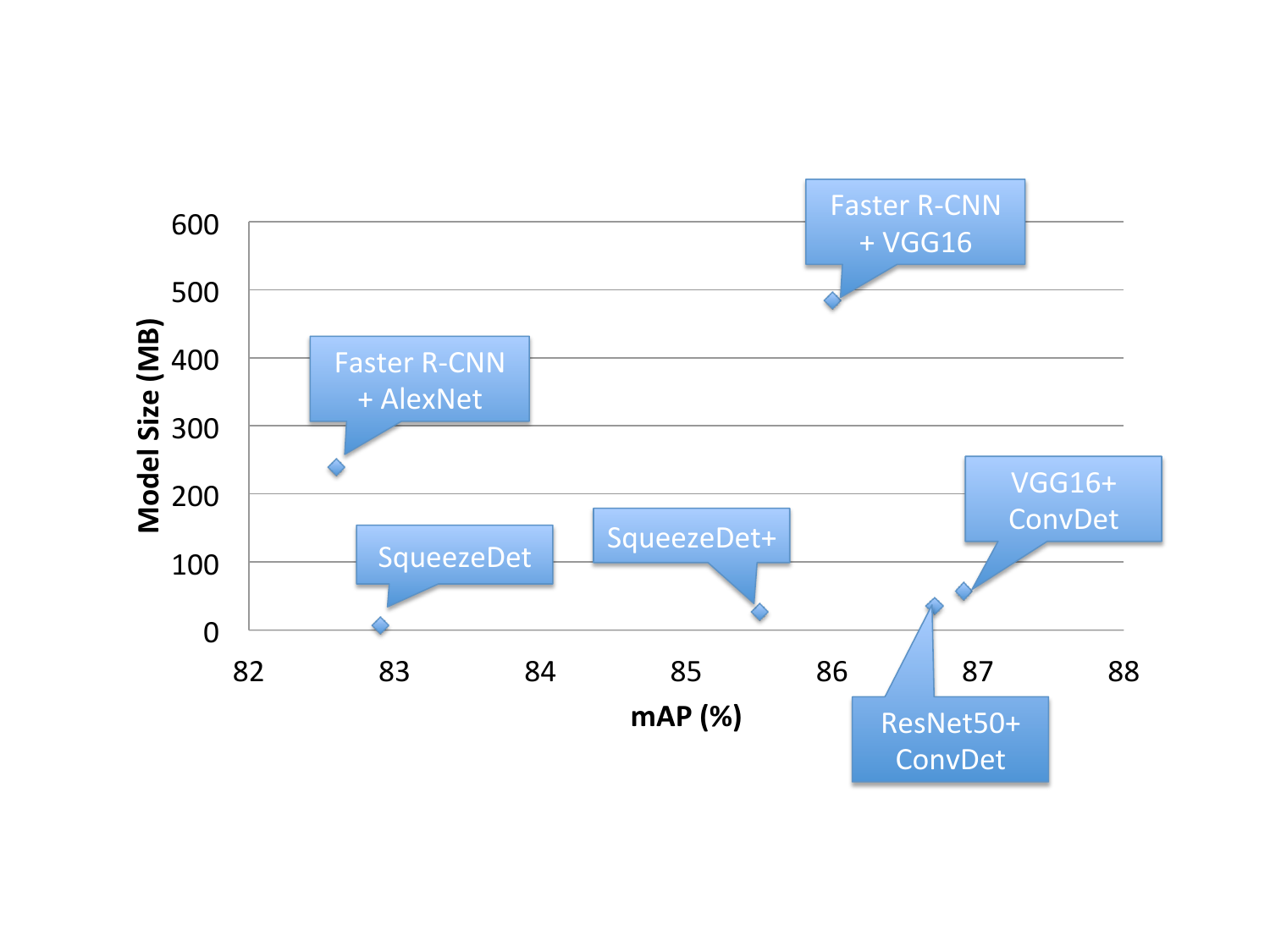}
\caption{Model size vs. mean average precision for car detection. Each point on this plane represents a method's model size and accuracy tradeoff.}
\label{fig:model-size-vs-mAP}
\end{figure}

\textbf{Model size.} We compare our proposed models with Faster-RCNN based models from~\cite{ShallowNetworks}. We plotted the model size and their mean average precision for three difficulty levels of the car category in Fig.~\ref{fig:model-size-vs-mAP} and summarized them in Table~\ref{table:AP}. As can be seen in Table~\ref{table:AP}, the SqueezeDet model is 61X smaller than the \textit{Faster R-CNN + VGG16} model, and it is 30X smaller than the \textit{Faster R-CNN + AlexNet} model. Almost $80\%$ of the parameters of the VGG16 model are from the fully connected layers. Thus, after we replace the fully connected layers and RPN layer with \textit{ConvDet}, the model size is only $57.4$MB. Compared with YOLO~\cite{YOLO} which is comprised of 24 convolutional layers, two fully connected layers with a model size of 753MB, SqueezeDet, without any compression, is 95X smaller. 

\subsection{Design space exploration} 
\vsp

\begin{table}
\begin{center}
\footnotesize
\begin{tabular}{c|ccccccc}
& &  & &   & Activation  &  \\
 & &  & & Model  & Memory  &  \\
 & mAP & Speed & FLOPs & Size & Footprint \\
DSE &  (\%) & (FPS) & $\times 10^9$  & (MB) & (MB) \\ 
\hline
SqueezeDet & 76.7 & 57.2  & 9.7 & 7.9 & 117.0 \\
scale-up & 72.4 & 31.3 & 22.5 & 7.9 & 263.3 \\
scale-down & 73.2 & 92.5 &5.3  & 7.9 & 65.8 \\
16 anchors & 66.9 & 51.4 & 11.0 & 9.4 & 117.4  \\
SqueezeDet+ & 80.4 & 32.1 & 77.2 & 26.8 & 252.7   \\
\hline
\end{tabular}
\end{center}
\caption{Design space exploration for SqueezeDet. Different approaches with their accuracy, FLOPs per image, inference speed, model size and activation memory footprint. The speed, FLOPS and activation memory footprint are measured for batch size of 1. We used mean average precision (mAP) to evaluate the overall accuracy on the KITTI object detection task. }
\label{table:DSE}
\end{table}

We conducted design space exploration to evaluate some key hyper-parameters' influence on our model's overall detection accuracy (measured in mAP). Meanwhile, we also investigated the ``cost'' of these variations in terms of FLOPs, inference speed, model size and memory footprint. The results are summarized in Table~\ref{table:DSE}, where the first row is our SqueezeDet architecture, subsequent rows are modifications to SqueezeDet, and the final row is SqueezeDet+.

\textbf{Image resolution.} For object detection, increasing image resolution is often an effective approach to improve detection accuracy~\cite{ShallowNetworks}. But, larger images lead to larger activations, more FLOPs, longer training time, etc. We now evaluate some of these tradeoffs. In our experiments, we scaled the image resolution by $1.5$X and $0.75$X receptively. With larger input image, the training becomes much slower, so we reduced the batch size to 10. 
As we can see in Table~\ref{table:DSE}, scaling up the input image actually decreases the mAP and also leads to more FLOPs, lower speed, and larger memory footprint. We also do an experiment with decreasing the image size. Scaling down the image leads to an astonishing 92.5 FPS of inference speed and a smaller memory footprint, though it suffers from a 3 percentage point drop in mean-average precision.

\textbf{Number of anchors.}
Another hyper-parameter to tune is the number of anchors. Intuitively, the more anchors to use, the more bounding box proposals are to be generated, thus should result in a better accuracy. However, in our experiment in Table~\ref{table:DSE}, using more anchors actually leads to lower accuracy. But, it also shows that for models that use \textit{ConvDet}, increasing the number of anchors only modestly increases the model size, FLOPs, and memory footprint. 

\textbf{Model architecture.}
As we discussed before, by using a more powerful backbone model with more parameters significantly improved accuracy (See Table~\ref{table:DSE}). But, this modification also costs substantially more in terms of FLOPs, model size and memory footprint.

\section{Conclusion}
\label{sec:conclusion}
\vsp

We proposed SqueezeDet, a fully convolutional neural network for real-time object detection. We integrated the region proposition and classification into \textit{ConvDet}, which is orders of magnitude smaller than its fully-connected counterpart. With the constraints of autonomous driving in mind, our proposed SqueezeDet and SqueezeDet+ models are designed to be small, fast, energy efficient, and accurate. Compared with previous baselines, we achieved the same acccuracy with 30.4x smaller model size, 19.7x faster inference speed, and 35.2x lower energy.

\clearpage
{\small
  \bibliographystyle{ieee}
  \bibliography{bibliography}
}

\end{document}


\title{Supplementary Material: Designing Low Power Neural Network Architectures 
}

\author{Bichen Wu$^1$, Forrest Iandola$^{1,2}$, Peter H. Jin$^1$, Kurt Keutzer$^{1,2}$ \\
UC Berkeley$^1$, DeepScale$^2$\\
{\tt\small bichen@berkeley.edu, forrest@deepscale.ai, phj@berkeley.edu, keutzer@berkeley.edu}
}

\maketitle

\section{Low Power Neural Net Design Guideline}
\vsp
Different operations involved in the computation of  a neural network consume different amounts of energy. According to~\cite{han2016eie}, a DRAM access consumes two orders of magnitude more energy than a SRAM access or a floating point arithmetic operation. In this work, our main focus is on reducing memory accesses. 

On-chip SRAM (Static Random Access Memory) and off-chip DRAM (Dynamic Random Access Memory) are the two major types of memory in computer hardware. Compared to off-chip DRAM, on-chip SRAM consumes about two orders of magnitude less energy, and SRAM read and write operations  have lower latency and higher bandwidth than accessing off-chip DRAM. However, SRAM requires more transistors to store the same amount of data compared to DRAM. Thus modern processors typically have a large off-chip DRAM-based main memory and a small  (\textit{e.g.} $16$MB) SRAM-based cache. During computation, processors prioritize SRAM for faster speed and lower energy consumption. But if the data size required for computation exceeds the on-chip SRAM capacity, processors will have to use off-chip DRAM. 

The degree to which programmers can control the utilization of on-chip SRAM versus off-chip DRAM depends considerably on the hardware. For example, GPU programming typically involves manual management of SRAM-based register files and shared memory \cite{nvidia_arch}. On ther other hand modern CPU processors are the results of decades of architecture research aimed at simplifying programming in general, and memory access in particular. As a result the programmer can typically only generally encourage cache locality by the structure of data access in the program, leaving the processor hardware to improve data locality through cache protocols and pre-fetching.
Thus, a simple and general rule to reduce energy consumed by memory accesses is to reduce the total memory footprint of the computations. In neural net computations this includes reducing the model parameters and intermediate layer activations. For hardware developers who aim to deploy the neural network on custom hardware (\textit{e.g.} on an FPGA), more granular memory scheduling becomes possible. With a neural net model with fewer model parameters and a perfect scheduling strategy, the hardware can cache all model parameters and activations of any two consecutive layers within on-chip SRAM, and no accesses to off-chip DRAM are necessary. This can lead to significant energy savings.

\section{Memory Footprint}
\label{sec:case_study}
\begin{figure}[h]
\centering
  \includegraphics[clip, trim=2cm 0cm 0cm 0cm, width=3.8in]{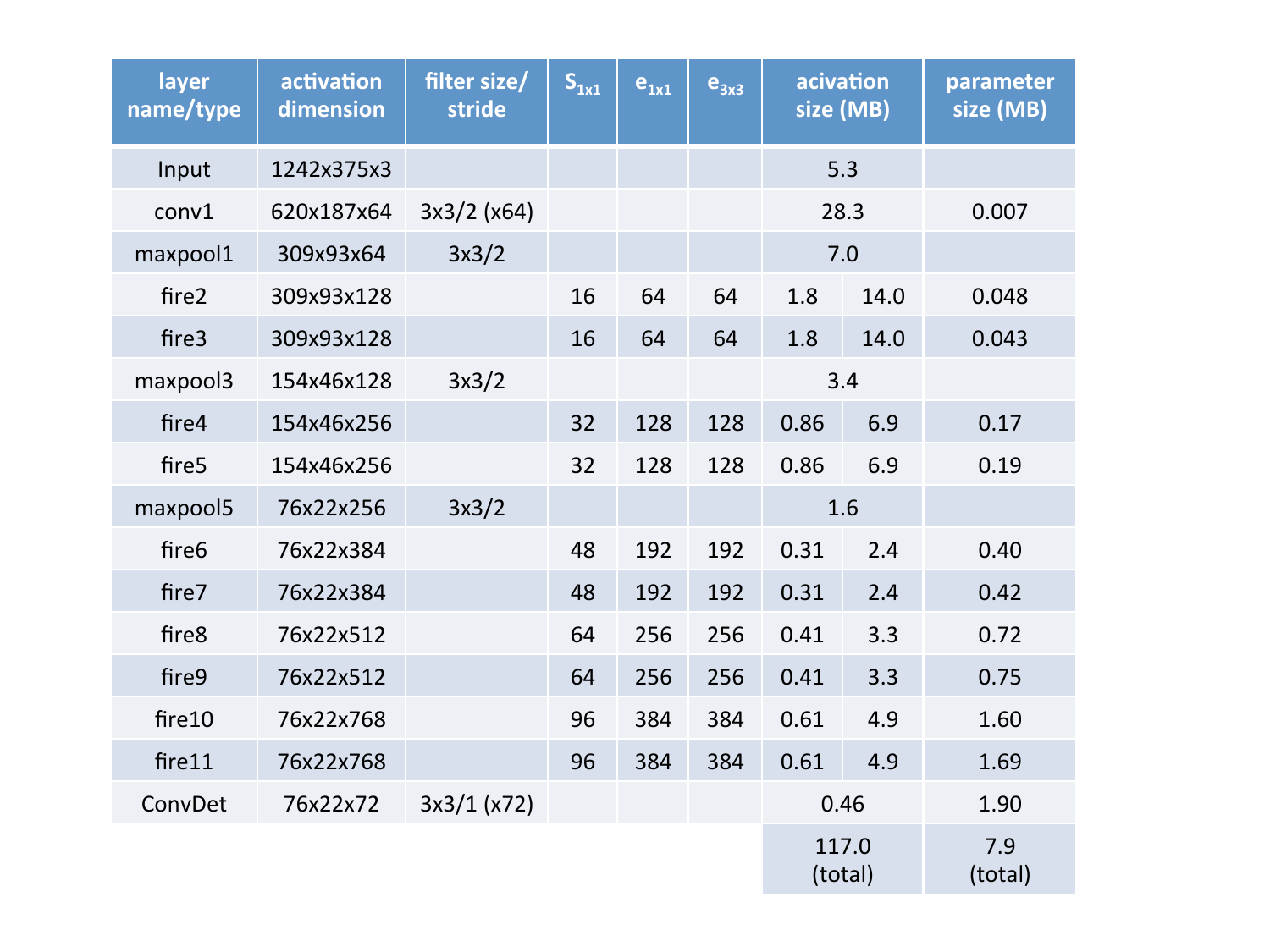}
  \captionof{table}{Layer specification of SqueezeDet. $s_{1x1}$ represents the number of 1x1 output filters in the \textit{squeeze} layer, $e_{1x1}$ is number of 1x1 filters in the \textit{expand} layer and $e_{3x3}$ is number of 3x3 filters in the \textit{expand} layer. 
 }
\label{fig:SqueezeDet} 
\end{figure}

\vsp
In what follows we analyze the memory footprint of SqueezeDet layer by layer. Details of the SqueezeDet model are shown in Table~\ref{fig:SqueezeDet}. The parameter size of SqueezeDet is just $7.9$MB without compression, so it's possible for many processors to fit the entire model in on-chip SRAM and reuse the parameters in evaluations. The largest intermediate activation is the output of the \textit{conv1} layer with $28.3$MB. \textit{conv1} is immediately followed by a max pooling layer. Potentially, we can fuse the implementation of max pooling and convolution layers such that the output of \textit{conv1} can be immediately down-sampled by 4X and we only need to store about $7$MB of activation to on-chip SRAM. Next, the \textit{maxpool1} output is fed into \textit{fire2}. The ``squeeze'' layer of the fire module compresses the input tensor and generates an activation with smaller channel size, and the two parallel ``expand'' layers of the fire module retrieve the compressed channel information and generate a larger output activation. The alternating ``squeeze'' and ``expand'' layers of the fire module effectively reduce the total size of activations of two consecutive layers. The following fire modules have increasingly larger channel size, but max pooling layers are used to reduce spatial resolution to control the activation size. Finally, even though the output of the final \textit{ConvDet} layer encodes thousands of bounding box proposals, its activation size is negligible.

We counted the activation memory footprint for several models, including SqueezeDet, variations thereof, and others. Our results are summarized in Table~\ref{table:energy}. As we can see, SqueezeDet has a much lower memory footprint and performs fewer FLOPs compared to other models, leading to better energy efficiency for SqueezeDet.

\begin{table*}
\begin{center}
\footnotesize
\begin{tabular}{c|ccccccc}
& & & Activation &  &  &  \\
& Model & & Memory & Average & Inference & Energy \\
& Size & FLOPs& Footprint &  GPU Power  &  Speed & Efficiency &mAP${}^{\star}$ \\ 
model & (MB) & $\times 10^9$ & (MB) &  (W)  &  (FPS) & (J/frame) &(\%) \\ 
\hline
SqueezeDet & 7.9 & 9.7 & 117.0 & 80.9 & 57.2 & 1.4 & 76.7 \\
SqueezeDet: scale-up & 7.9 & 22.5 & 263.3 & 89.9 & 31.3 & 2.9 & 72.4 \\
SqueezeDet: scale-down & 7.9 & 5.3 & 65.8 & 77.8 & 92.5 & 0.84 & 73.2 \\
SqueezeDet: 16 anchors & 9.4 & 11.0 & 117.4 & 82.9 & 51.4 & 1.6& 66.9\\
SqueezeDet+ & 26.8 & 77.2 & 252.7 & 128.3 & 32.1 & 4.0 & 80.4 \\
\hline
VGG16+ConvDet & 57.4 & 288.4 & 540.4 & 153.9 &16.6 & 9.3 & 79.1 \\
ResNet50+ConvDet & 35.1 & 61.3 & 369.0  & 95.4 &22.5 & 4.2 & 76.1 \\
\hline
\hline
Faster-RCNN + VGG16~\cite{ShallowNetworks} & 485 &- & - & 200.1 & 1.7 & 117.7 & -\\
Faster-RCNN + AlexNet~\cite{ShallowNetworks} & 240 & - & - & 143.1 & 2.9& 49.3 & -\\
YOLO${}^{\star\star}$ & 753 & - & - & 187.3 & 25.8 & 7.3 & -\\
\end{tabular}
\end{center}
\caption{Comparing SqueezeDet and other models in terms of Energy efficiency and other aspects. The default image resolution is 1242x375, but the ``SqueezeDet: scale-up'' variation up-sampled input image's height and width by 1.5X. The ``scale-down'' variation scaled image resolution by 0.75X. The default SqueezeDet model contains 9 anchors. But the 16-anchor variation contains 16 anchors for each grid. 
${}^{\star}$ The mAP denotes the mean average precision of 3 difficulty levels of 3 categories on KITTI dataset. It represents each model's detection accuracy on KITTI dataset. ${}^{\star\star}$ We launched YOLO to detect  $4,952$ VOC 2007 test images and it took 192 seconds to finish. We then compute the inference speed as $4,952/192 \approx$ $25.8$FPS, which is slower than the speed reported in~\cite{YOLO}. The input image to YOLO is scaled to 448x448.}
\label{table:energy}
\end{table*}

\section{Experiments}
\label{sec:Experiments}
\begin{figure}[h]
\centering
  \includegraphics[clip, trim=0cm 0cm 0cm 0cm, width=3in]{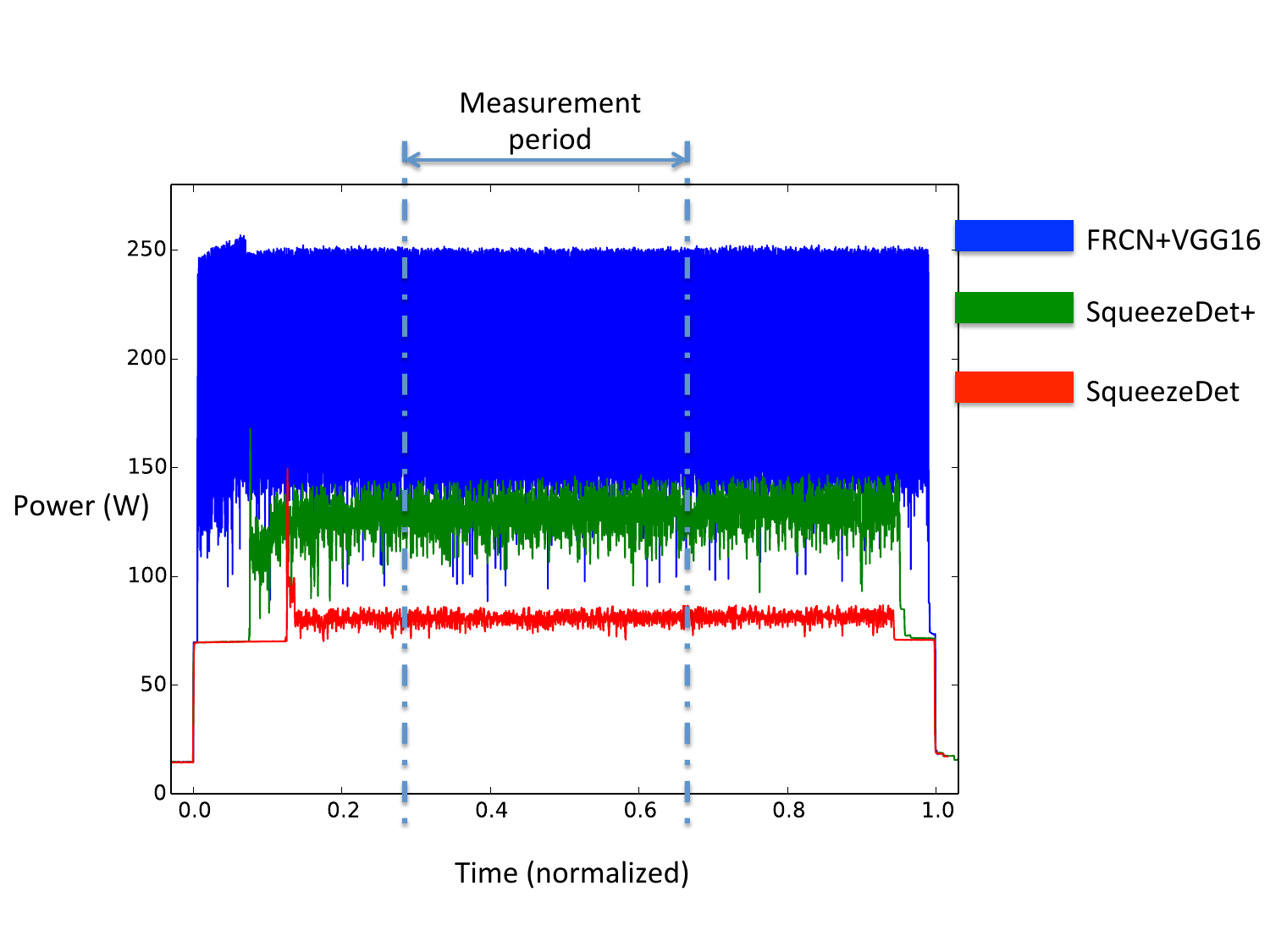}
  \caption{GPU power measured by \texttt{nvidia-smi}. Here we plot power measurement curve of 3 models, SqueezeDet, SqueezeDet+, and Faster R-CNN + VGG16 model. We normalize the working period of 3 models to the same range of [0, 1]. We divide the working period evenly into 3 parts and use the middle part to compute the average GPU power for each model.}
\label{fig:power-curve}
\end{figure}

\vsp
We measured the energy consumption of SqueezeDet and the other models during the object detection evaluation of $3741$ images from the KITTI dataset~\cite{KITTI}. The default input image resolution is 1242x375, and the batch size is set to 1. Meanwhile, we measured the GPU power usage with Nvidia's system monitor interface (\texttt{nvidia-smi}).  We sampled the power reading with a fixed interval of 0.1 second. Then, we obtained the power-vs-time curve as shown in Fig~\ref{fig:power-curve}.  When the GPU is idle, it consumes about $15$W of power. Through the evaluation process, the GPU went through several stages from idle to working and then to idle again. We denote the period with power measurement $\ge20$W as working period. Then, we divide the working period evenly into 3 parts, and we take the measurements from the middle part to compute the average GPU power. The energy consumption per image is then computed as \[\frac{\text{Average Power [Joule/Second]}}{\text{Inference Speed [Frame/Second]}}.\]

We measured energy consumption of SqueezeDet and several other models using the above approach, and our experimental results are listed in Table~\ref{table:energy}.  SqueezeDet consumes only $1.4$J per image, which is 84$\times$ less than the Faster R-CNN + VGG16 model.  Scaling the image resolution down by 0.75$\times$, the mAP drops by 3 percentage points, but the inference speed is 1.6$\times$ faster and the energy consumption is less than $1$J per image. With much better accuracy, SqueezeDet+  only consumes $4$J per image, which is $>$10X more efficient than Faster R-CNN based methods. We combined the convolutional layers of VGG16 and ResNet50 with ConvDet, both models achieved much better energy efficiency compared with Faster R-CNN based models.

We also compared our models with YOLO. We use YOLO to detect $4,952$ images from the VOC 2007~\cite{pascal-voc-2007} test set. The input images are scaled to 448x448, batch size is 1. It took YOLO $192$ seconds to finish the evaluation. Using the same approach to measure the GPU power of YOLO, we compute the energy per frame of YOLO as 7.3J. Using the frame rate of 45FPS which is reported in ~\cite{YOLO}, YOLO's energy consumption per frame is 4.2J, which is comparable with SqueezeDet+. But note that input image (with size of 1242x375) to SqueezeDet+ in our experiment contains 2X more pixels than the input image (448x448) to YOLO.  

Our experiments show that SqueezeDet and its variations are very energy efficient compared with previous neural network based object detectors.

{\small
  \bibliographystyle{ieee}
  \bibliography{bibliography}
}